\documentclass[journal,twoside,web]{ieeecolor}
\usepackage[table]{xcolor}
\usepackage{tmi}
\usepackage{cite}
\usepackage{amsmath,amssymb,amsfonts}
\usepackage{algorithmic}
\usepackage{graphicx}
\usepackage{textcomp}
\usepackage{multirow}
\usepackage{makecell}
\usepackage{booktabs}
\usepackage{threeparttable}
\usepackage{arydshln}
\usepackage[ruled,vlined]{algorithm2e}

\def\BibTeX{{\rm B\kern-.05em{\sc i\kern-.025em b}\kern-.08em
    T\kern-.1667em\lower.7ex\hbox{E}\kern-.125emX}}
\begin{document}

\title{RADA: Region-Aware Dual-encoder Auxiliary Learning for Barely-supervised Medical Image Segmentation}
\author{Shuang Zeng, Boxu Xie, Lei Zhu, Xinliang Zhang, JiaKui Hu, Zhengjian Yao, Yuanwei Li, Yuxing Lu, Yanye Lu
}

\maketitle

\begin{abstract}


Deep learning has greatly advanced medical image segmentation, but its success relies heavily on fully supervised learning, which requires dense annotations that are 
costly and time-consuming for 3D volumetric scans. Barely-supervised learning reduces annotation burden by using only a few labeled slices per volume. Existing methods typically propagate sparse annotations to unlabeled slices through geometric continuity to generate pseudo-labels, but this strategy lacks semantic understanding, often resulting in low-quality pseudo-labels. Furthermore, medical image segmentation is inherently a pixel-level visual understanding task, where accuracy fundamentally depends on the quality of local, fine-grained visual features. Inspired by this, we propose RADA, a novel Region-Aware Dual-encoder Auxiliary learning pipeline which introduces a dual-encoder framework pre-trained on Alpha-CLIP to extract fine-grained, region-specific visual features from the original images and limited annotations. The framework combines image-level fine-grained visual features with text-level semantic guidance, providing region-aware semantic supervision that bridges image-level semantics and pixel-level segmentation. Integrated into a triple-view training framework, RADA achieves SOTA performance under extremely sparse annotation settings on LA2018, KiTS19 and LiTS, demonstrating robust generalization across diverse datasets. 
\end{abstract}


\section{Introduction}
\label{sec:introduction}
\IEEEPARstart{D}{eep} learning has significantly advanced medical image segmentation \cite{c3loss,AttUKAN}, enabling precise delineation of anatomical structures and pathological regions crucial for diagnosis, treatment planning, and disease monitoring. However, these achievements predominantly rely on {\itshape fully supervised learning (FSL)}, which demands large-scale, pixel-wise annotated datasets. For volumetric modalities such as CT and MRI, which often comprise hundreds of slices per scan, it is extremely time-consuming and costly to obtain the annotations, posing a major barrier to clinical deployment.

\begin{figure}[t]
  \centering
  \includegraphics[width=0.5\textwidth]{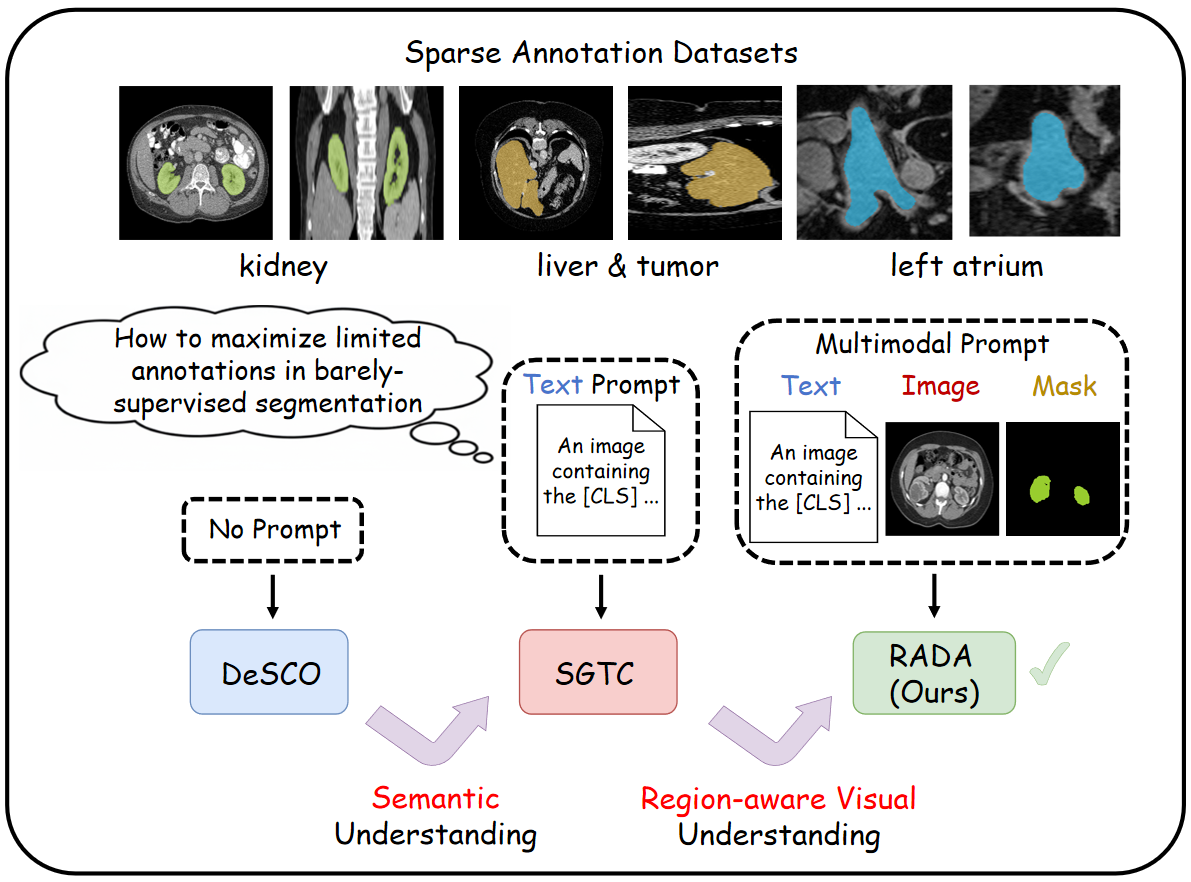}
  \caption{Conceptual comparison of different prompt-based learning paradigms for barely-supervised medical image segmentation. The diagram illustrates three distinct approaches applied to sparse annotation datasets. From left to right: (1) DeSCO (No Prompt) relies solely on geometric propagation without semantic guidance; (2) SGTC (Text Prompt) incorporates textual semantics through class-specific prompts; (3) Our proposed RADA (Multimodal Prompt) integrates complementary modalities -- combining text prompts with original image data and sparse annotation masks to achieve region-aware visual understanding.}
  \label{intro}
\end{figure}

To reduce annotation costs, {\itshape semi-supervised learning (SSL)} \cite{MT,UA-MT,CPS,SuperCL,MACL} leverages both limited labeled data and abundant unlabeled samples. While SSL alleviates the labeling burden, it still requires fully annotated slices for part of the dataset, which remains impractical for large 3D volumes. This limitation has motivated {\itshape barely-supervised learning (BSL)} \cite{PLN,DeSCO,SGTC}, an even more annotation-efficient paradigm where only a few slices per volume are labeled, accompanied by numerous unlabeled volumes. By drastically minimizing annotation demands, BSL offers a clinically feasible path toward scalable medical segmentation.

However, the central challenge of BSL lies in its extremely sparse supervision -- only a few labeled slices are available to guide the model. 
Consequently, the key question becomes how to {\itshape maximally exploit these limited supervision signals} 
({\itshape i.e.}, sparse annotations) to effectively drive learning. Existing BSL methods such as DeSCO \cite{DeSCO} 
address this challenge by generating pseudo-labels through inter-slice registration, propagating limited annotations 
to neighboring slices. These methods leverage geometric continuity to expand supervision but rely solely on intensity 
and spatial correspondence, {\itshape lacking semantic understanding}. As a result, the generated pseudo-labels often 
degrade in regions with weak boundaries or subtle intensity variations, leading to blurred or inconsistent segmentation 
of complex anatomical structures. On the other hand, medical image segmentation is 
{\itshape inherently a pixel-level visual understanding task}, where segmentation accuracy fundamentally depends on the 
quality of local, fine-grained feature representations. Therefore, it becomes crucial to extract 
{\itshape rich, region-specific visual features} from the limited annotations and available original images to provide 
stronger pixel-level supervision and more discriminative feature learning.

Motivated by these insights, also shown in Figure \ref{intro}, we propose RADA (Region-Aware Dual-encoder Auxiliary learning) -- a novel framework that enables {\itshape region-aware, fine-grained semantic guidance} for barely-supervised medical image segmentation. Specifically, RADA introduces sparse annotation masks as auxiliary spatial guidance alongside original RGB images, enabling the model to focus explicitly on clinically relevant regions. To fully exploit this auxiliary mask information, we introduce a region-aware encoder that deeply integrates the sparse annotations with visual features, learning local, fine-grained, and semantically aligned representations that maximize the utility of scarce supervision in BSL settings.
Furthermore, we develop a dual-encoder collaborative framework that synergistically combines visual and textual semantic guidance. The region-aware visual encoder provides semantically aligned features with precise spatial localization derived from the auxiliary masks, while the text encoder delivers explicit category-level semantic guidance. Through hierarchical cross-modal fusion, RADA effectively combines these complementary cues to achieve rich, region-aware semantic supervision for pixel-level segmentation. Built upon this region-aware dual-encoder foundation, RADA is incorporated into a unified triple-view training framework for 3D medical image segmentation. Extensive experiments on LA2018, KiTS19, and LiTS datasets demonstrate that under extremely sparse annotation settings -- only three orthogonal slices per volume -- RADA achieves state-of-the-art performance, substantially outperforming existing methods and showing strong generalization across diverse anatomical structures and imaging modalities.

\section{Related Works}

\subsection{Semi-supervised Medical Image Segmentation}

Recently, semi-supervised learning methods have become increasingly popular in medical image segmentation, as they can effectively leverage both small amounts of labeled data and large volumes of unlabeled data. Existing semi-supervised methods can be broadly classified into two categories. The first category comprises pseudo-label-based methods, such as Mean Teacher (MT) \cite{MT} and Uncertainty-Aware Mean Teacher (UA-MT) \cite{UA-MT}, which estimate pseudo labels based on a few labeled samples. Building upon this foundation, BCP \cite{BCP} employs a bidirectional copy-paste strategy to reduce the distribution gap between labeled and unlabeled data. The second category encompasses consistency-based methods, including SASSNet \cite{SASSNet}, Cross Pseudo Supervision (CPS) \cite{CPS}, and DTC \cite{DTC}, which leverage consistency regularization across different sub-networks. Despite their effectiveness, these approaches still require full annotation of volumetric samples, which limits their applicability in clinical practice.

Therefore, barely-supervised (or sparsely annotated semi-supervised) segmentation have been developed to address the annotation burden by utilizing only a few representative 2D annotated slices within each 3D volume. 
Early approaches focus on label propagation through geometric or spatial correspondence. For instance, DeSCO \cite{DeSCO} utilizes annotations from two orthogonal slices and employs image registration method to construct cross-view pseudo-labels, thereby expanding the supervision signal. While these registration-based methods significantly reduce annotation costs, they fundamentally lack high-level semantic understanding. A significant advancement comes with SGTC \cite{SGTC}, which introduces a semantic-guided CLIP-based framework and leverages sparse annotations from three orthogonal slices.  
While SGTC demonstrated the value of semantic guidance, its use of standard CLIP provides only a global, coarse-grained semantic prior. This guidance is limited to the text encoder, lacking the fine-grained, region-aware visual features for pixel-level segmentation task.

\begin{figure*}[t]
  \centering
  \includegraphics[width=\textwidth]{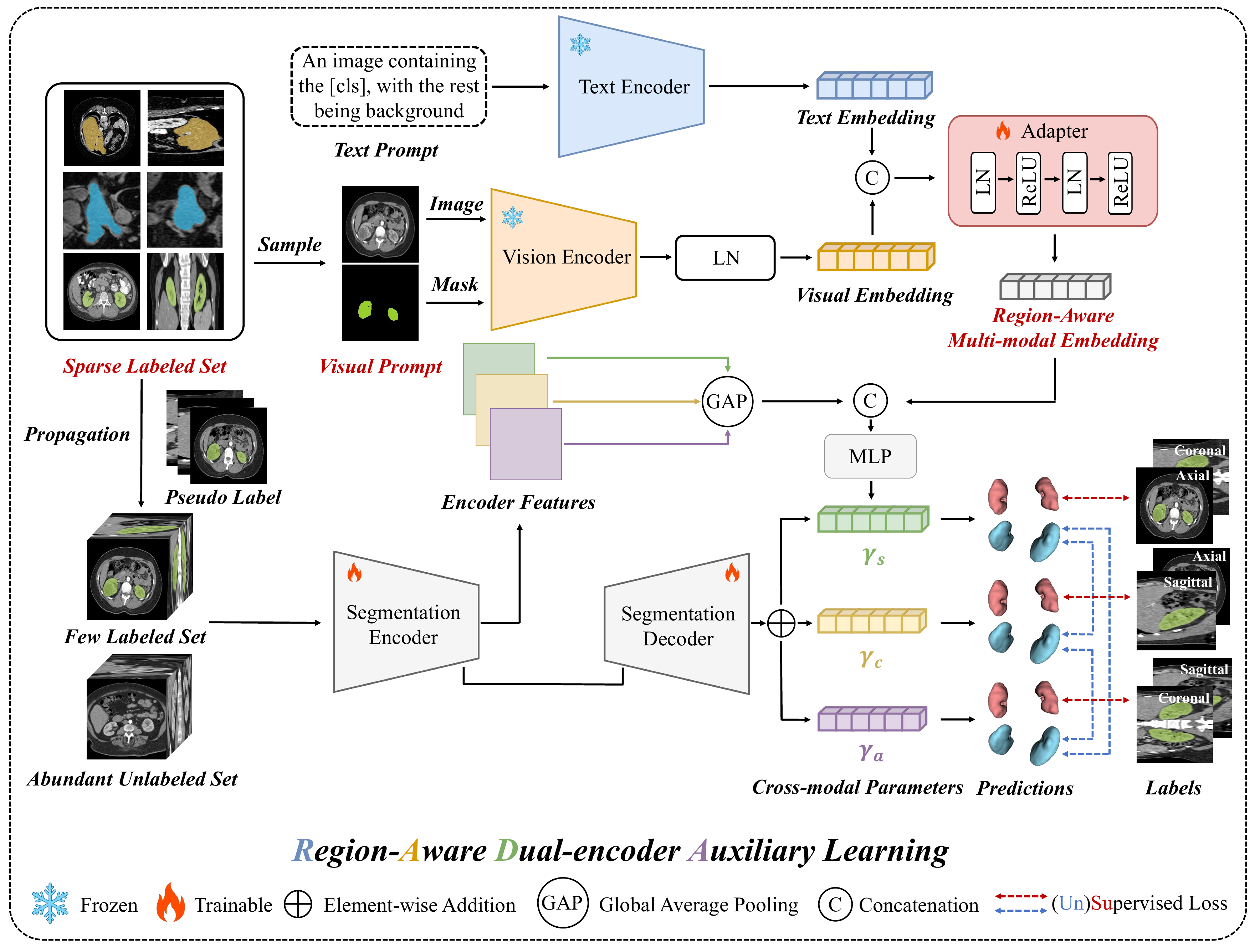}
  \caption{The pipeline of our proposed RADA framework. Specifically, a frozen vision encoder extracts region-aware visual features from the paired original images and their corresponding masks while simultaneously a frozen text encoder processes medical-domain prompts to generate semantic embeddings, with both modalities fused through an Adapter module for cross-modal alignment. The segmentation stage employs three parallel 3D sub-networks corresponding to orthogonal views, each incorporating the fused features via auxiliary learning to enhance decoder predictions through cross-modal guidance. The final output generates refined segmentation masks through coordinated multi-view predictions, utilizing registration-based pseudo-labels to expand limited supervision during training, thereby effectively leveraging sparse annotations through region-aware feature learning and multi-modal integration for precise segmentation.}
  \label{RADA}
\end{figure*}

\subsection{Region-Aware Vision-Language Models}

Vision-Language Models (VLMs) such as CLIP \cite{CLIP} align text embeddings with global image representations but lack the ability to focus on specific local regions, which is critical for segmentation tasks demanding fine-grained anatomical understanding. To address this limitation, visual prompting has emerged as a powerful solution to inject explicit spatial information and guide the model's attention toward regions of interest. Recent efforts have explored incorporating pixel-level visual information into CLIP. MaskCLIP \cite{MaskCLIP} introduces masked self-distillation into contrastive language-image pretraining, enabling the model to learn fine-grained local semantics with indirect language supervision while maintaining global vision-language alignment. Alpha-CLIP \cite{Alpha-CLIP} extends the standard CLIP architecture by incorporating an additional alpha channel to process visual prompts, enabling the model to generate region-aware features while preserving global semantic understanding. PixCLIP \cite{PixCLIP} achieves fine-grained visual language understanding by aligning pixel-level visual features with text tokens at any granularity, enabling dense pixel-text correspondence learning beyond the conventional image-level contrastive pretraining. In this paper, we attempt to integrate a region-aware visual encoder into barely-supervised segmentation pipeline so that we can fully exploit the supervision information from sparse annotations. 
\section{Methodology}

\subsection*{Preliminaries}
In this work, we define a training dataset $D$ that consists of a labeled subset $\Omega_L$ and an unlabeled subset $\Omega_U$. The labeled set $\Omega_L = \{(X_i^l, Y_i)\}_{i=1}^M$ contains $M$ samples, where $X_i^l$ denotes the input image and $Y_i$ represents the corresponding label. The unlabeled set $D_U = \{X_i^u\}_{i=1}^N$ includes $N$ unlabeled samples, with $N \gg M$. For barely-supervised learning setting, we will select three representative slices from three orthogonal planes (sagittal, coronal and axial) with their corresponding labels for supervision. The three representative slices denote as $X_{is}^{lp}$, $X_{ic}^{lq}$, and $X_{ia}^{lr}$, respectively. Here, $p$, $q$, and $r$ refer to the indices of the selected slices in each plane. The corresponding labels of these slices are expressed as $Y_{is}^p$, $Y_{ic}^q$, and $Y_{ia}^r$, respectively.

\subsection{Region-Aware Dual-encoder Auxiliary Learning}

Figure \ref{RADA} illustrates our proposed Region-Aware Dual-encoder Auxiliary learning (RADA) framework. Our proposed RADA addresses the core challenge of barely-supervised learning -- extremely sparse annotations -- by integrating original medical images and sparse annotation masks into a unified dual-encoder framework. Traditional methods ({\itshape e.g.}, DeSCO) rely solely on geometric propagation, which lacks semantic understanding and struggles with complex anatomical boundaries. In contrast, RADA leverages the original image to capture global contextual features (textures, shapes) and the sparse mask as a visual prompt to explicitly guide the model’s focus to clinically relevant regions. This combination enables the model to extract region-specific visual features that are fine-grained and semantically aligned. The mask constrains the feature learning process to foreground structures, reducing ambiguity in regions with weak boundaries or low contrast. Simultaneously, the text encoder processes domain-specific prompts ({\itshape e.g., ``An image containing the [CLS]”}) to provide categorical semantic guidance. The visual and textual representations are fused via an Adapter module, bridging pixel-level details with high-level semantics. Then the fused features are injected into three parallel 3D segmentation sub-networks (sagittal, coronal, axial) through an auxiliary learning mechanism. This design ensures that each view-specific decoder benefits from both localized visual cues and global semantic context, significantly improving segmentation accuracy under extreme annotation sparsity. By unifying geometric propagation (via registration-based pseudo-labels) and semantic guidance, our RADA maximizes the utility of limited supervision while maintaining robustness across diverse anatomical structures.

\subsubsection{Region-Aware Dual-encoder}
Specifically, our RADA employs a dual-encoder design to extract both semantic and visual representations. For the text pathway, we utilize the pre-trained text encoder from Alpha-CLIP to process medical prompts formulated as {\itshape ``An image containing the [CLS], with the rest being background"}, where the text prompt is originated from SGTC and [CLS] represents the specific anatomical structure of interest ({\itshape e.g. kidney, liver}). The text embedding ${E_t}$ is computed as:  

\begin{equation}
  E_t = e_t(\text{Text Prompt})
\end{equation}
where $e_t$ is the text encoder of Alpha-CLIP. 

For the visual pathway, we employ the pre-trained vision encoder from Alpha-CLIP to extract region-aware visual representations from the input image $X_{ia}^{lr}$ (take the axial labeled slice as an example) and its corresponding mask $Y_{ia}^r$. These features are transformed through a linear projection to obtain the visual embedding $E_v$:

\begin{equation}
  E_v = e_v(X_{ia}^{ir}, Y_{ia}^r)
\end{equation}
where $e_v$ is the vision encoder of Alpha-CLIP.
Given the inherent domain gap between natural images (on which Alpha-CLIP was pretrained) and medical images, directly applying Alpha-CLIP-derived features may limit the model's ability to capture subtle clinical semantics. Therefore, we freeze the pre-trained Alpha-CLIP encoders and introduce a lightweight trainable Adapter module for cross-modal domain adaptation. The Adapter module consists of linear (LN) transformation and ReLU activation, specifically structured as ``LN $\rightarrow$ ReLU $\rightarrow$ LN $\rightarrow$ ReLU". Finally, we can get the region-aware multi-modal embedding $E_{ra}$:

\begin{equation}
  E_{ra} = \text{Adapter}(E_t \text{ \copyright} \ E_v)
\end{equation}
where \text{\copyright} denotes concatenation. This design enables effective alignment between semantic and visual modalities while adapting the pre-trained representations to the medical imaging domain.

\begin{table*}[htbp]
\centering
\caption{Quantitative comparisons with 8 SOTA methods on LA2018 dataset under 10\% labeled cases. \textbf{bold} values denote the best performace. S, C, A indicate sagittal, coronal and axial annotated slices.}
\label{LA_table}
\setlength{\tabcolsep}{10pt} 
 \scriptsize 
\begin{tabular}{l c cc cccc} 
\toprule
\multirow{2}{*}{\textbf{Method}} & \multirow{2}{*}{\textbf{Labeled Slices}} & \multicolumn{2}{c}{\textbf{Scans Used}} & \multicolumn{4}{c}{\textbf{Metrics}} \\
\cmidrule(lr){3-4} \cmidrule(lr){5-8} 
& & \textbf{Labeled} & \textbf{Unlabeled} & \textbf{Dice $\uparrow$} & \textbf{Jaccard $\uparrow$} & \textbf{HD $\downarrow$} & \textbf{ASD $\downarrow$} \\
\midrule
MT \cite{MT} & SA & 8 & 72 & 0.7365 $\pm$ 0.0801 & 0.5895 $\pm$ 0.1039 & 36.79 $\pm$ 10.09 & 9.42 $\pm$ 3.07 \\
UA-MT \cite{UA-MT} & SA & 8 & 72 & 0.6615 $\pm$ 0.0912 & 0.5011 $\pm$ 0.1055 & 44.27 $\pm$ 7.11 & 17.66 $\pm$ 3.84\\
SASSNet \cite{SASSNet} & SA & 8 & 72 & 0.7767 $\pm$ 0.1582 & 0.6532 $\pm$ 0.1501 & 23.73 $\pm$ 13.22 & 5.70 $\pm$ 3.19\\
CPS \cite{CPS} & SA & 8 & 72 & 0.7859 $\pm$ 0.0976 & 0.6571 $\pm$ 0.1219 & 27.44 $\pm$ 9.48 & 6.39 $\pm$ 2.53 \\
DTC \cite{DTC} & SA & 8 & 72 & 0.7975 $\pm$ 0.1452 & 0.6806 $\pm$ 0.1526 & 17.84 $\pm$ 8.89 & \textbf{2.69 $\pm$ 1.32} \\
BCP \cite{BCP} & SA & 8 & 72 & 0.8084 $\pm$ 0.1506 & 0.6969 $\pm$ 0.1552 & 16.53 $\pm$ 10.19 & 3.19 $\pm$ 1.36\\
DeSCO \cite{DeSCO} & SA & 8 & 72 & 0.7189 $\pm$ 0.0827 & 0.5673 $\pm$ 0.1012 & 35.63 $\pm$ 7.87 & 12.13 $\pm$ 4.17 \\
SGTC \cite{SGTC} & SCA & 8 & 72 & 0.8429 $\pm$ 0.0444 & 0.7309 $\pm$ 0.0647 & 19.91 $\pm$ 13.92 & 4.42 $\pm$ 3.66\\
RADA (Ours) & SCA & 8 & 72 & \textbf{0.8652 $\pm$ 0.0282} & \textbf{0.7634 $\pm$ 0.0441} & \textbf{15.17 $\pm$ 10.90} & {3.24 $\pm$ 2.23}\\
\bottomrule
\end{tabular}
\end{table*}

\subsubsection{Auxiliary Learning for Segmentation}

The segmentation branch consists of three parallel 3D segmentation sub-networks, denoted as $F^{s}(\cdot)$, $F^{c}(\cdot)$, and $F^{a}(\cdot)$, corresponding to sagittal, coronal, and axial views, respectively. Each sub-network adopts an encoder-decoder architecture augmented with our cross-modal guidance mechanism for auxiliary learning.
For each sub-network, visual features extracted from the encoder are first aggregated via global average pooling (GAP) to get the pooled features $f_{img}$. Then the pooled features are concatenated with the region-aware multi-modal embedding $E_{ra}$ and subsequently fed into a multi-layer perceptron (MLP) to obtain view-specific cross-modal parameters $\gamma_{s/c/a}$ for each sub-network:
\begin{equation}
\gamma_{s/c/a} = \text{MLP}(E_{ra} \text{ \copyright}\ f_{img}^{s/c/a})  
\end{equation}

To incorporate cross-modal guidance for auxiliary learning, we perform element-wise addition between the cross-modal parameters and 
the features ${Z}_{s/c/a}$ extracted before the final classification layer of each 
decoder, and pass them through a 3D convolution layer to obtain the final 
predictions:

\begin{equation}
P_{s/c/a} = \text{Conv3D}({Z}_{s/c/a} \oplus {\gamma}_{s/c/a}),
\end{equation}
where $\oplus$ denotes element-wise addition, and ${P}_{s}$, ${P}_{c}$, ${P}_{a}$ are the segmentation predictions for sagittal, coronal, and axial views, respectively.

\subsubsection{Training Strategy}

\begin{table*}[htbp]
\centering
\caption{Quantitative comparisons with 8 SOTA methods on KiTS19 dataset under 10\% labeled cases.}
\label{KiTS_table}
\setlength{\tabcolsep}{10pt} 
\scriptsize 
\begin{tabular}{l c cc cccc} 
\toprule
\multirow{2}{*}{\textbf{Method}} & \multirow{2}{*}{\textbf{Labeled Slices}} & \multicolumn{2}{c}{\textbf{Scans Used}} & \multicolumn{4}{c}{\textbf{Metrics}} \\
\cmidrule(lr){3-4} \cmidrule(lr){5-8} 
& & \textbf{Labeled} & \textbf{Unlabeled} & \textbf{Dice $\uparrow$} & \textbf{Jaccard $\uparrow$} & \textbf{HD $\downarrow$} & \textbf{ASD $\downarrow$} \\
\midrule
MT \cite{MT} & SA & 19 & 171 & 0.8324 $\pm$ 0.1022 & 0.7255 $\pm$ 0.1438 & 31.86 $\pm$ 13.92 & 6.79 $\pm$ 4.48 \\
UA-MT \cite{UA-MT} & SA & 19 & 171 & 0.8303 $\pm$ 0.0717 & 0.7160 $\pm$ 0.1052 & 30.58 $\pm$ 7.78 & 9.38 $\pm$ 3.41\\
SASSNet \cite{SASSNet} & SA & 19 & 171 & 0.8737 $\pm$ 0.0362 & 0.7775 $\pm$ 0.0574 & 23.70 $\pm$ 5.88 & 6.33 $\pm$ 1.79\\
CPS \cite{CPS} & SA & 19 & 171 & 0.8335 $\pm$ 0.0510 & 0.7177 $\pm$ 0.0722 & 47.24 $\pm$ 7.02 & 12.51 $\pm$ 2.51 \\
DTC \cite{DTC} & SA & 19 & 171 &  0.6747 $\pm$ 0.0507 & 0.5114 $\pm$ 0.0615 & 63.33 $\pm$ 6.55 & 0.51 $\pm$ 0.19\\
BCP \cite{BCP} & SA & 19 & 171 &  0.6725 $\pm$ 0.0612 & 0.5097 $\pm$ 0.0781 & 63.06 $\pm$ 6.52 & 0.58 $\pm$ 0.39 \\
DeSCO \cite{DeSCO} & SA & 19 & 171 & 0.8873 $\pm$ 0.0527 & 0.8011 $\pm$ 0.0834 & 8.15 $\pm$ 5.96 & 2.50 $\pm$ 1.31\\
SGTC \cite{SGTC} & SCA & 19 & 171 & 0.9254 $\pm$ 0.0349 & 0.8629 $\pm$ 0.0589 & 8.54 $\pm$ 11.33 & 2.56 $\pm$ 2.46\\
RADA (Ours) & SCA & 19 & 171 & \textbf{0.9448 $\pm$ 0.0138} & \textbf{0.8957 $\pm$ 0.0245} & \textbf{3.09 $\pm$ 2.43} & \textbf{1.30 $\pm$ 0.98}\\
\bottomrule 
\end{tabular}
\end{table*}

\begin{table*}[htbp]
\centering
\caption{Quantitative comparisons with 8 SOTA methods on LiTS dataset under 10\% labeled cases.}
\label{LiTS_table}
\setlength{\tabcolsep}{10pt} 
\scriptsize 
\begin{tabular}{l c cc cccc} 
\toprule
\multirow{2}{*}{\textbf{Method}} & \multirow{2}{*}{\textbf{Labeled Slices}} & \multicolumn{2}{c}{\textbf{Scans Used}} & \multicolumn{4}{c}{\textbf{Metrics}} \\
\cmidrule(lr){3-4} \cmidrule(lr){5-8} 
& & \textbf{Labeled} & \textbf{Unlabeled} & \textbf{Dice $\uparrow$} & \textbf{Jaccard $\uparrow$} & \textbf{HD $\downarrow$} & \textbf{ASD $\downarrow$} \\
\midrule
MT \cite{MT} & SA & 10 & 90 & 0.8990 $\pm$ 0.0341 & 0.8183 $\pm$ 0.0564 & 21.94 $\pm$ 12.64 & 3.73 $\pm$ 1.43 \\
{UA-MT} \cite{UA-MT} &SA&10 &90 &0.6336 $\pm$ 0.0677&0.4671 $\pm$ 0.0706&61.82 $\pm$ 11.60&25.52 $\pm$ 5.92\\
{SASSNet} \cite{SASSNet} &SA &10&90&0.6408 $\pm$ 0.0701&0.4751 $\pm$ 0.0738&76.94 $\pm$ 13.55&31.70 $\pm$ 5.99\\
CPS \cite{CPS} &SA & 10 & 90 & 0.8935 $\pm$ 0.0494 & 0.8110 $\pm$ 0.0768 & 10.04 $\pm$ 8.37 & 2.18 $\pm$ 1.38 \\
DTC \cite{DTC} &SA &10&90&0.9206 $\pm$ 0.0408&0.8554 $\pm$ 0.0673&6.26 $\pm$ 5.75&1.65 $\pm$ 1.32\\
{BCP} \cite{BCP} &SA &10&90&0.9244 $\pm$ 0.0498&0.8629$\pm$ 0.0788&8.81 $\pm$ 11.21&2.53 $\pm$ 2.45 \\
DeSCO \cite{DeSCO} &SA &10&90&0.8400 $\pm$  0.0908  &  0.7334 $\pm$ 0.1219 & 20.84 $\pm$ 13.00 & 4.68 $\pm$ 2.87\\
SGTC \cite{SGTC} &SCA &10&90& 0.9256 $\pm$ 0.0337 & 0.8631 $\pm$ 0.0548 & 10.08 $\pm$ 10.67 & 3.02 $\pm$ 2.19\\
RADA (Ours) &SCA&10&90&\textbf{0.9363 $\pm$ 0.0203} & \textbf{0.8809 $\pm$ 0.0348} & \textbf{5.45 $\pm$ 6.53} & \textbf{1.78 $\pm$ 1.49}\\
\bottomrule
\end{tabular}
\end{table*}

Following the triple-view training strategy of SGTC, our framework leverages three orthogonal annotated slices to supervise the three sub-networks while maintaining their disparity and enabling complementary learning. Building upon this foundation, we introduce an additional regularization mechanism through a registration-based pseudo-label generation approach inspired by DeSCO. For each volume ${X}^{l}$ with sparse orthogonal annotations, we employ a registration module $M_{\text{reg}}$ to perform slice-by-slice label propagation from the annotated slices to generate dense pseudo labels $\hat{{Y}}_{s}$, $\hat{{Y}}_{c}$, and $\hat{{Y}}_{a}$ for each view. To account for error accumulation in label propagation, we assign voxel-wise weights based on the distance from the registration source slice. The weight for voxel $i$ in view $a$ is defined as:
\begin{equation}
W_{a}^{i} = \begin{cases}
1, & \text{if voxel } i \text{ is labeled in } Y \\
\alpha^{d}, & \text{otherwise},
\end{cases}
\end{equation}
where $d$ is the distance from voxel $i$ to the registration source slice and $\alpha \in [0, 1)$ is the decay rate. We then apply label mixing to combine the sparse annotations with the registration-based pseudo labels, producing mixed pseudo labels $\tilde{{Y}}_{s/c/a}$ with corresponding weight maps ${W}_{s/c/a}$.

For the supervised loss on labeled data, we employ a combination of weighted cross-entropy loss and weighted Dice loss:
\begin{equation}
\mathcal{L}_{\text{WCE}} = -\frac{1}{\sum_{i=1}^{H \times W \times D} w_{i}} \sum_{i=1}^{H \times W \times D} w_{i} y_{i} \log p_{i},
\end{equation}
\begin{equation}
\mathcal{L}_{\text{Dice}} = 1 - \frac{2 \times \sum_{i=1}^{H \times W \times D} w_{i} p_{i} y_{i}}{\sum_{i=1}^{H \times W \times D} w_{i}(p_{i}^{2} + y_{i}^{2})},
\end{equation}
where $w_{i}$ is the $i^{th}$ voxel value of the weight matrix ${W}_{s/c/a}$, $p_{i}$ and $y_{i}$ respectively represent the predicted results and the labels on voxel $i$. The supervised loss is:
\begin{equation}
\mathcal{L}_{\text{Sup}} = \frac{1}{2} \mathcal{L}_{\text{WCE}} + \frac{1}{2} \mathcal{L}_{\text{Dice}}.
\end{equation}


For unlabeled volumes, each sub-network's predictions serve as cross-supervision signals for the other two sub-networks. The unsupervised loss for each sub-network $k$ is computed as:
\begin{equation}
\mathcal{L}_{\text{Unsup}}^{k} = -\frac{1}{\sum_{i=1}^{H \times W \times D} m_{i}} \sum_{i=1}^{H \times W \times D} m_{i} \hat{y}_{i}^{k} \log p_{i}, \quad k = 1, 2
\end{equation}
where $m_{i}$ denotes whether the $i^{th}$ voxel is selected based on uncertainty estimation, $p_{i}$ is the prediction result of the current sub-network, and $\hat{y}_{i}^{1}$ and $\hat{y}_{i}^{2}$ are the pseudo labels generated by the other two sub-networks.

The overall loss function during training is defined as:
\begin{equation}\label{total}
\mathcal{L}_{\text{total}} = (1 - \beta) \mathcal{L}_{\text{Sup}} + \beta \mathcal{L}_{\text{Unsup}},
\end{equation}
where $\beta$ is a dynamic parameter that gradually increases during training to balance supervised and unsupervised learning. This design enables effective learning from limited orthogonal annotations while maintaining high segmentation accuracy across diverse medical imaging scenarios.

\section{Experimental Results}

\subsection{Datasets}

We evaluate our method on three public medical image segmentation benchmarks:
(1) LA2018 dataset \cite{xiong2021global} comprises 100 gadolinium-enhanced MRI scans of the left atrium with corresponding segmentation labels. All scans have an isotropic resolution of $0.625 \times 0.625 \times 0.625$ mm$^3$. Following prior works \cite{UA-MT}, we adopt the standard split of 80 training samples and 20 testing samples to ensure fair comparison.
(2) KiTS19 dataset \cite{heller2019kits19} contains 300 abdominal CT scans collected from the Medical Centre of Minnesota University for kidney and kidney tumor segmentation. The slice thickness varies from 1mm to 5mm. We utilize 190 scans for training and 20 scans for testing.
(3) LiTS dataset \cite{bilic2023liver} focuses on liver and liver tumor segmentation from CT images, comprising 201 abdominal scans. Among these, 131 scans with segmentation masks are publicly available. Following the split used in \cite{cai2023orthogonal}, we employ 100 scans for training and 31 scans for testing.

\begin{table*}[t]
\centering
\scriptsize
\setlength{\tabcolsep}{5pt}
\caption{Ablation Study of each component on LA2018 dataset.}
\label{ablation_LA}
\begin{tabular}{cccccccc}
\toprule
\multirow[]{2}{*}{\textbf{Method}} & \multicolumn{3}{c}{\textbf{Settings}} & \multicolumn{4}{c}{\textbf{Metrics}} \\
\cmidrule(lr){2-4} \cmidrule(lr){5-8}
 & {\itshape Image Prompt} & {\itshape Mask Prompt} & {\itshape Registration} & \textbf{Dice $\uparrow$} & \textbf{Jaccard $\uparrow$} & \textbf{HD $\downarrow$} & \textbf{ASD $\downarrow$} \\
\midrule
SGTC (Baseline) & $\times$ & $\times$ & $\times$ & 0.8429 $\pm$ 0.0444 & 0.7309 $\pm$ 0.0647 & 19.91 $\pm$ 13.92 & 4.42 $\pm$ 3.66 \\
SGTC + CLIP Vision Encoder & $\checkmark$ & $\times$ & $\times$ & 0.8504 $\pm$ 0.0382 & 0.7416 $\pm$ 0.0578 & 17.24 $\pm$ 12.44 & 4.43 $\pm$ 3.66 \\
SGTC + Alpha-CLIP Vision Encoder & $\checkmark$ & $\checkmark$ & $\times$ & 0.8600 $\pm$ 0.0368 & 0.7561 $\pm$ 0.0561 & 18.23 $\pm$ 13.80 & 4.55 $\pm$ 3.92 \\
RADA (Ours) & $\checkmark$ & $\checkmark$ & $\checkmark$ & \textbf{0.8652 $\pm$ 0.0282} & \textbf{0.7634 $\pm$ 0.0441} & \textbf{15.17 $\pm$ 10.90} & \textbf{3.24 $\pm$ 2.23} \\
\bottomrule
\end{tabular}
\end{table*}

\begin{table*}[htbp]
\centering
\scriptsize
\setlength{\tabcolsep}{5pt}
\caption{Ablation Study of each component on KiTS19 dataset.}
\label{ablation_KiTS}
\begin{tabular}{cccccccc}
\toprule
\multirow{2}{*}{\textbf{Method}} & \multicolumn{3}{c}{\textbf{Settings}} & \multicolumn{4}{c}{\textbf{Metrics}} \\
\cmidrule(lr){2-4} \cmidrule(lr){5-8}
 & {\itshape Image Prompt} & {\itshape Mask Prompt} & {\itshape Registration} & \textbf{Dice $\uparrow$} & \textbf{Jaccard $\uparrow$} & \textbf{HD $\downarrow$} & \textbf{ASD $\downarrow$} \\
\midrule
SGTC (Baseline) & $\times$ & $\times$ & $\times$ & 0.9254 $\pm$ 0.0349 & 0.8629 $\pm$ 0.0589 & 8.54 $\pm$ 11.33 & 2.56 $\pm$ 2.46 \\
SGTC + CLIP Vision Encoder & $\checkmark$ & $\times$ & $\times$ & 0.9381 $\pm$ 0.0269 & 0.8846 $\pm$ 0.0464 & 10.08 $\pm$ 14.92 & 2.67 $\pm$ 2.48 \\
SGTC + Alpha-CLIP Vision Encoder & $\checkmark$ & $\checkmark$ & $\times$ & 0.9392 $\pm$ 0.0241 & 0.8862 $\pm$ 0.0420 & 4.59 $\pm$ 8.04 & 1.58 $\pm$ 1.54 \\
RADA (Ours) & $\checkmark$ & $\checkmark$ & $\checkmark$ & \textbf{0.9448 $\pm$ 0.0138} & \textbf{0.8957 $\pm$ 0.0245} & \textbf{3.09 $\pm$ 2.43} & \textbf{1.30 $\pm$ 0.98} \\
\bottomrule
\end{tabular}
\end{table*}

\begin{table*}[htbp]
\centering
\scriptsize
\setlength{\tabcolsep}{5pt}
\caption{Ablation Study of each component on LITS dataset.}
\label{ablation_LiTS}
\begin{tabular}{cccccccc}
\toprule
\multirow{2}{*}{\textbf{Method}} & \multicolumn{3}{c}{\textbf{Settings}} & \multicolumn{4}{c}{\textbf{Metrics}} \\
\cmidrule(lr){2-4} \cmidrule(lr){5-8}
 & {\itshape Image Prompt} & {\itshape Mask Prompt} & {\itshape Registration} & \textbf{Dice $\uparrow$} & \textbf{Jaccard $\uparrow$} & \textbf{HD $\downarrow$} & \textbf{ASD $\downarrow$} \\
\midrule
SGTC (Baseline) & $\times$ & $\times$ & $\times$ & 0.9256 $\pm$ 0.0337 & 0.8631 $\pm$ 0.0548 & 10.08 $\pm$ 10.67 & 3.02 $\pm$ 2.19 \\
SGTC + CLIP Vision Encoder & $\checkmark$ & $\times$ & $\times$ & 0.9309 $\pm$ 0.0337 & 0.8725 $\pm$ 0.0557 & 8.66 $\pm$ 11.29 & 2.61 $\pm$ 2.45 \\
SGTC + Alpha-CLIP Vision Encoder & $\checkmark$ & $\checkmark$ & $\times$ & 0.9319 $\pm$ 0.0312 & 0.8739 $\pm$ 0.0526 & 7.74 $\pm$ 7.82 & 2.39 $\pm$ 1.88 \\
RADA (Ours) & $\checkmark$ & $\checkmark$ & $\checkmark$ & \textbf{0.9363 $\pm$ 0.0203} & \textbf{0.8809 $\pm$ 0.0348} & \textbf{5.45 $\pm$ 6.53} & \textbf{1.78 $\pm$ 1.49} \\
\bottomrule
\end{tabular}
\end{table*}

\begin{table*}[htbp]
\centering
\caption{Ablation Study of text prompts in Alpha-CLIP on LA2018, KiTS19 and LiTS dataset.}
\label{text_prompt}
\renewcommand{\arraystretch}{1.3}
\setlength\dashlinedash{3pt}
\setlength\dashlinegap{3pt}
\scriptsize
\setlength{\tabcolsep}{5pt}
\begin{tabular}{>{\centering\arraybackslash}p{3.5cm}cccccc}
\toprule
\multirow{2}{*}{\textbf{Text Prompts}} & \multicolumn{2}{c}{\textbf{LA2018}} & \multicolumn{2}{c}{\textbf{KiTS19}} & \multicolumn{2}{c}{\textbf{LiTS}} \\
\cmidrule(lr){2-3} \cmidrule(lr){4-5} \cmidrule(lr){6-7}
 & \textbf{Dice $\uparrow$} & \textbf{Jaccard $\uparrow$} & \textbf{Dice $\uparrow$} & \textbf{Jaccard $\uparrow$} & \textbf{Dice $\uparrow$} & \textbf{Jaccard $\uparrow$} \\
\midrule
None. & 0.8573 $\pm$ 0.0313 & 0.7515 $\pm$ 0.0479 & 0.9401 $\pm$ 0.0138 & 0.8872 $\pm$ 0.0244 & 0.9302 $\pm$ 0.0246 & 0.8705 $\pm$ 0.0417 \\
\hdashline
A photo of a [cls]. & 0.8518 $\pm$ 0.0376 & 0.7435 $\pm$ 0.0553 & 0.9408 $\pm$ 0.0159 & 0.8887 $\pm$ 0.0283 & 0.9336 $\pm$ 0.0217 & 0.8761 $\pm$ 0.0371 \\
\hdashline
\makecell{There is a [cls] in this \\ computerized tomography/ \\ magnetic resonance imaging.} & 0.8554 $\pm$ 0.0298 & 0.7485 $\pm$ 0.0462 & 0.9409 $\pm$ 0.0149 & 0.8890 $\pm$ 0.0264 & 0.9346 $\pm$ 0.0209 & 0.8780 $\pm$ 0.0358 \\
\hdashline
\makecell{An image containing the [cls], \\ with the rest being background.} & \textbf{0.8652 $\pm$ 0.0282} & \textbf{0.7634 $\pm$ 0.0441} & \textbf{0.9448 $\pm$ 0.0138} & \textbf{0.8957 $\pm$ 0.0245} & \textbf{0.9363 $\pm$ 0.0203} & \textbf{0.8809 $\pm$ 0.0348} \\
\bottomrule
\end{tabular}
\end{table*}

\begin{figure}[h]
  \centering
  \includegraphics[width=0.5\textwidth]{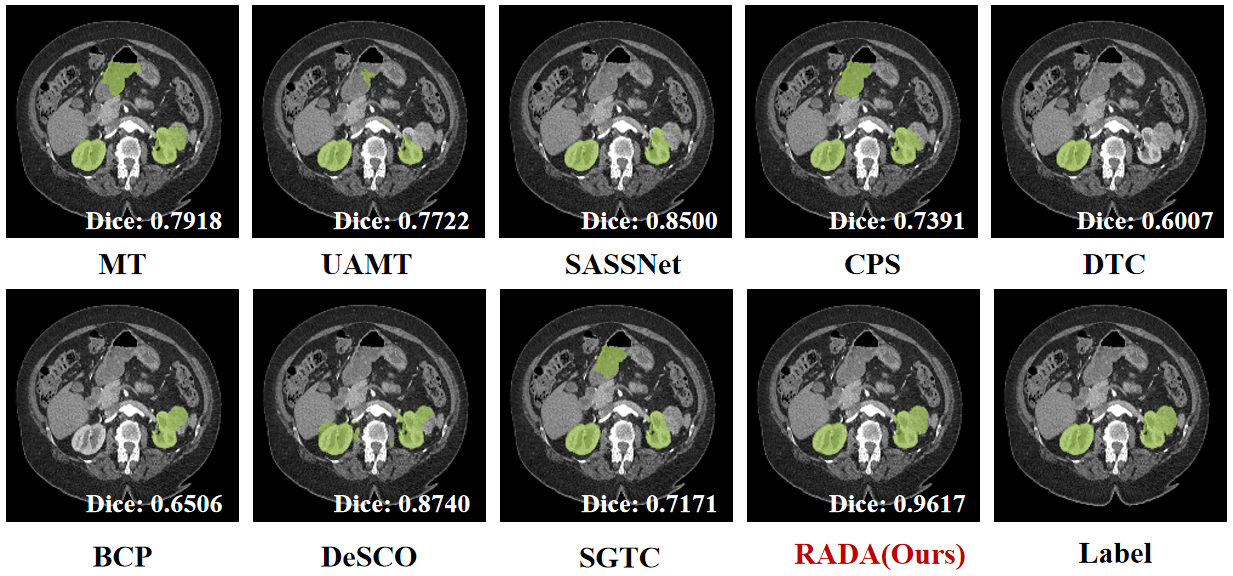}
  \caption{Quantitative comparisons on KiTS19 dataset with 10\% labeled case.}
  \label{LA_seg}
\end{figure}

\begin{figure}[h]
  \centering
  \includegraphics[width=0.5\textwidth]{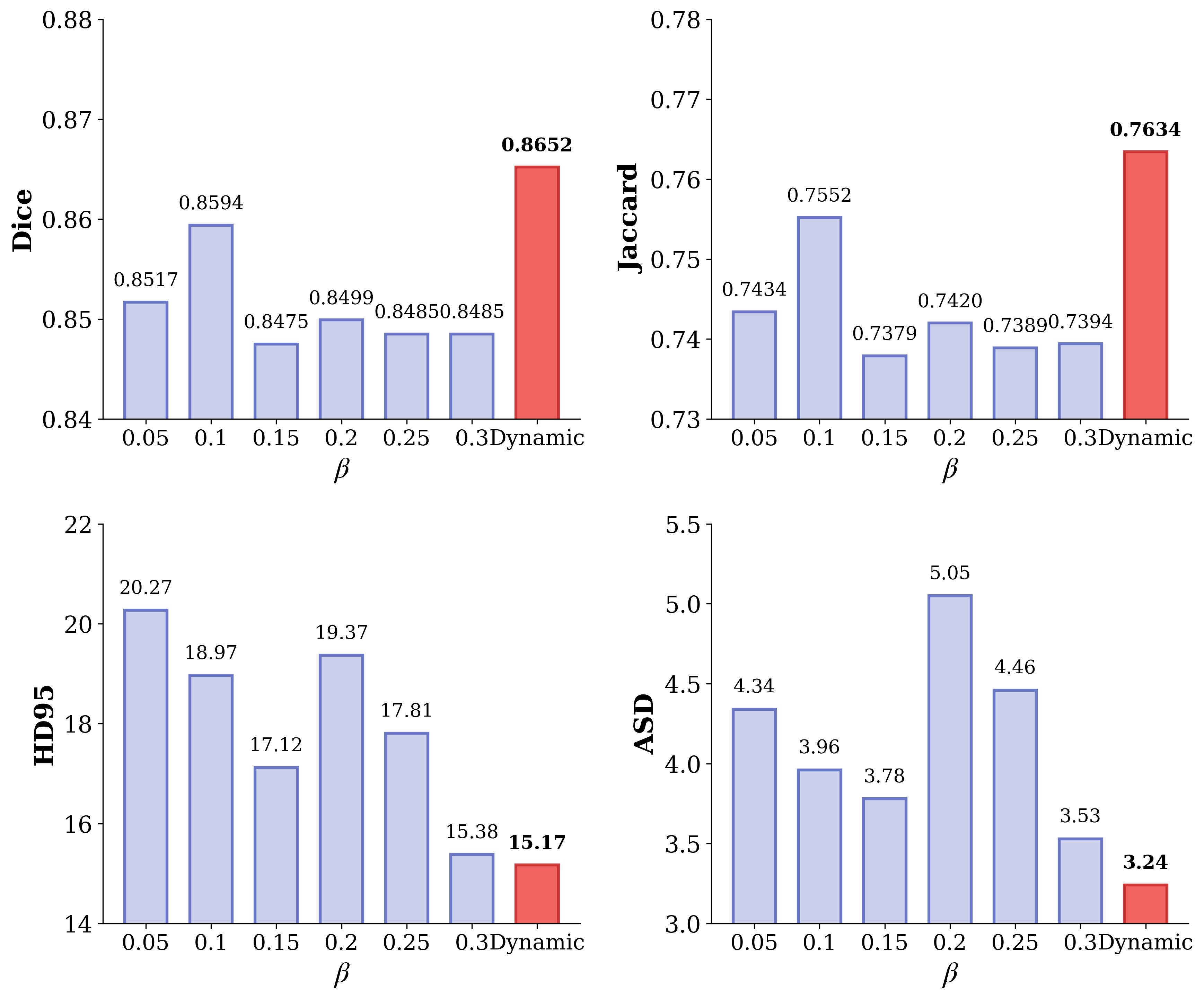}
  \caption{Analysis of dynamic parameter $\beta$.}
  \label{beta}
\end{figure}

\subsection{Implementation Details}
For sparse supervision, we select three orthogonal slices, prioritizing those that capture a large foreground area of the target organ. For the implementation of our method, we adopt V-Net \cite{vnet} as the backbone of our segmentation model and SyNRA in Ants \cite{Ants} as our registration module. We set the total batch size to 4, which is composed of 2 labeled volumes and 2 unlabeled volumes. We employ the Stochastic Gradient Descent (SGD) optimizer for all three networks. The learning rate is initialized at 0.01 and followed by a step decay schedule, which is reduced by a factor of 0.1 every 2500 iterations, eventually decaying down to 0.0001. The slice-wise weight decay rate $\alpha$ is initialized to 0.97 and update every 1000 iterations according to cosine rampdown until it decays to 0. The dynamic parameter $\beta$ is initialized to 0.1 and increases every 150 iterations. Experiments are conducted with PyTorch on NVIDIA L40S 48G GPUs. For fair comparison with other methods, we use four commonly used segmentation metrics: Dice, Jaccard, 95\% Hausdorff Distance (HD95), and Average Surface Distance (ASD).

\subsection{Comparison Experiments}

We evaluate our RADA against 6 methods for semi-supervised medical image segmentation: MT \cite{MT}, UA-MT \cite{UA-MT}, SASSNet \cite{SASSNet}, CPS \cite{CPS}, DTC \cite{DTC}, BCP \cite{BCP}, and 2 methods specifically designed for barely-supervised medical image segmentation: DeSCO \cite{DeSCO} and SGTC \cite{SGTC}. Experiments are conducted on three challenging benchmarks: LA2018, KiTS19 and LiTS. Following prior work \cite{DeSCO}, we annotate three slices (sagittal, coronal and axial plane) per volume for SGTC and two slices (sagittal and axial plane) per volume for other baseline methods. 

According to the quantitative results of Table \ref{LA_table}, \ref{KiTS_table} and \ref{LiTS_table}, our proposed RADA demonstrates consistent state-of-the-art performance across all three medical image segmentation benchmarks under the challenging 10\% labeled data setting. Specifically, on LA2018, RADA achieves the highest Dice of 0.8652 and Jaccard of 0.7634, outperforming SGTC by 2.23\% of Dice and 3.25\% of Jaccard. For KiTS19, our RADA outperforms the baseline methods in all metrics, it reaches 0.9448 Dice with significantly improved boundary detection (HD 3.09). On LiTS, RADA maintains leadership with top Dice of 0.9363 and Jaccard of 0.8809. Its superior performance on boundary metrics: 5.45 of HD and 1.78 of ASD further highlights its capability for accurate boundary delineation. Reasonably, our proposed RADA's region-aware dual-encoder auxiliary learning framework effectively leverages extremely sparse annotations to extract rich, region-specific visual features and semantic guidance. This enables precise voxel-level segmentation and robust generalization across diverse organs and complex boundaries, establishing SOTA performance in barely-supervised medical image segmentation. Moreover, visual comparisons are shown in Figure \ref{LA_seg}, where our DARA accurately delineates the intricate structure and weak boundaries with the highest Dice. More visualizations can be found in the supplementary materials.    

\subsection{Ablation Study and Analysis}

\subsubsection{Ablation study of each component}

The ablation studies presented in Tables \ref{ablation_LA}, \ref{ablation_KiTS} and \ref{ablation_LiTS} systematically validate the contribution of each component in our RADA framework across three medical image segmentation benchmarks. The results demonstrate a clear, progressive improvement in performance with each added component, confirming our design rationale for addressing the challenges of barely-supervised learning. Starting from the SGTC baseline (Dice: 0.8429 on LA2018), incorporating the CLIP vision encoder yields an initial gain (0.8504) by leveraging powerful, pre-trained visual features that enhance semantic understanding. Then a more significant improvement is achieved by replacing it with the Alpha-CLIP vision encoder (0.8600), where the key advancement is the use of a mask prompt. This component allows the model to ingest the sparse annotation masks via an auxiliary alpha channel, directing its focus explicitly to anatomical regions of interest and enabling the extraction of fine-grained, region-aware features crucial for boundary delineation. 
Finally, with the integration of the registration module to propagate sparse labels to neighboring slices, our RADA pushes its performance to its peak (0.8652). This stepwise enhancement -- from generic visual features to region-specific guidance and finally to denser pseudo-supervision—systematically addresses the core challenge of barely-supervised learning, confirming that each component works synergistically to maximize the utility of extremely sparse annotations.

\subsubsection{Parameter Analysis}

Figure \ref{beta} evaluates the hyper-parameter $\beta$ in Eq. \ref{total}, Specifically, we have conducted ablation experiments on LA2018 dataset by setting $\beta$ to [0.05, 0.1, 0.15, 0.2, 0.25, 0.3] and the dynamic parameter. Compared with the fixed values, while $\beta=0.1$ achieves the best fixed performance (Dice: 0.8594), the dynamic parameter setting yields superior results (Dice: 0.8652) by adaptively balancing supervised and unsupervised losses during training. This demonstrates that dynamic adjustment enhances training stability and segmentation accuracy under sparse annotations.

\subsubsection{Ablation study of text prompts in Alpha-CLIP}

Table \ref{text_prompt} demonstrates that incorporating textual guidance significantly enhances segmentation performance across all datasets. While the baseline without text prompts achieves solid results (0.8573 Dice on LA2018), the tailored prompt {\itshape ``An image containing the [CLS], with the rest being background"} consistently yields the best performance (0.8652 Dice on LA2018, 0.9448 on KiTS19). This optimized formulation explicitly defines the foreground-background relationship, providing clearer semantic guidance that helps distinguish target structures from surrounding tissues.

\section{Conclusion}

In this paper, we proposed RADA, a novel Region-Aware Dual-encoder Auxiliary learning framework to tackle the critical challenge of extremely sparse supervision in barely-supervised medical image segmentation. Our key innovation lies in the introduction of a region-aware dual-encoder architecture that strategically leverages sparse annotation masks as explicit visual prompts, enabling the model to focus on clinically relevant regions and extract fine-grained, discriminative features from the original images. This approach effectively bridges the gap between image-level semantics and pixel-level segmentation.
Extensive experiments on LA2018, KiTS19, and LiTS datasets demonstrate that under extremely sparse annotation settings -- only three orthogonal slices per volume -- RADA achieves SOTA performance,substantially outperforming existing methods and showing strong generalization across diverse anatomical structures and imaging modalities.

\bibliographystyle{IEEEtran}
\bibliography{tmi.bib}

\newpage
\section{Appendix}

\subsection{Network Architecture}

The backbone network employed in our RADA is VNet, which is a 3D fully convolutional encoder-decoder network specifically designed for volumetric medical image segmentation, following a symmetric U-shaped architecture with skip connections to preserve fine-grained spatial information. 

\textbf{Encoder Path:} The encoder consists of five hierarchical stages that progressively extract features at multiple scales. The feature channels increase from 16 to 32, 64, 128, and finally 256 as the network goes deeper. Each stage contains convolutional blocks composed of $3 \times 3 \times 3$ convolutions followed by BatchNorm3d normalization and ReLU activation. The depth of convolutional layers varies across stages: the first stage uses a single convolutional layer, the second stage employs two layers, while stages three through five each contain three convolutional layers to capture increasingly complex patterns. Between consecutive stages, downsampling is performed using $2 \times 2 \times 2$ strided convolutions, which reduce the spatial resolution by half while doubling the number of feature channels.

\textbf{Decoder Path:} The decoder symmetrically mirrors the encoder structure through four upsampling stages, progressively recovering spatial resolution while reducing feature channels ($256 \to 128 \to 64 \to 32 \to 16$). Each upsampling operation is implemented using $2 \times 2 \times 2$ transposed convolutions followed by BatchNorm3d and ReLU activation. Skip connections play a crucial role by concatenating feature maps from the corresponding encoder stages with the upsampled decoder features, enabling the network to combine high-level semantic information with low-level spatial details. Each decoder stage also contains convolutional blocks that refine the fused features before passing them to the next stage.

\textbf{Output Head:} The final decoder stage produces a 16-channel feature map, which is then processed by a $1 \times 1 \times 1$ convolution to generate the 2-class segmentation output. A dropout layer with probability 0.5 is incorporated for regularization during training.

The multi-layer perceptron (MLP), mentioned in Equation (4) to generate cross-modal parameters $\gamma_{s/c/a}$, is implemented as a sequence of four $1 \times 1 \times 1$ Conv3d layers. It performs a progressive channel reduction on the input features, mapping 512 channels to 8 channels ($512 \to 256 \to 128 \to 64 \to 8$).

\subsection{Pseudo Code}
The overall training pipeline of our proposed RADA framework is summarized in Algorithm \ref{alg:rada_pipeline}. First, we generate dense pseudo-labels from sparse annotations using registration to create a better supervision signal. Second, the Region-Aware Dual-encoder extracts a unified feature embedding by combining the region-aware visual features and text embedding. Finally, this embedding guides the learning process, and the model is updated by minimizing both supervised and unsupervised losses.

\begin{algorithm*}[t]
\caption{Training Pipeline of Our RADA Framework}
\label{alg:rada_pipeline}
\SetKwInput{KwInput}{Input}
\SetKwInput{KwOutput}{Output}

\KwInput{Labeled set $\Omega_L$; Unlabeled set $\Omega_U$; Pre-trained Alpha-CLIP image encoders $e_v$ and text encoders $e_t$.}
\KwOutput{Optimized sub-networks $F^s, F^c, F^a$.}

\For{$iter \leftarrow 1$ \KwTo $MaxIterations$}{
    Sample batch $(X^l, Y^l) \sim \Omega_L$ and $X^u \sim \Omega_U$\;
    
    Obtain mixed pseudo labels ${Y}_{\text{s/c/a}}$ and weights $W$ via registration-based propagation\;
    
    $E_t \leftarrow e_t(\text{``An image containing the {[}CLS{]}, with the rest being background''})$\;
    $E_{\text{ra}} \leftarrow \text{Adapter}(E_t \mathbin{\copyright} e_v(X^l, Y^l)))$\;
    
    \For{$k \in \{s, c, a\}$}{
        Extract image features $f^k_{\text{img}}$ from encoder of $F^k$\;
        $\gamma_k \leftarrow \text{MLP}(E_{\text{ra}} \mathbin{\copyright} \text{GAP}(f^k_{\text{img}}))$\;
        $P_k \leftarrow \text{Conv3D}(Z_k \oplus \gamma_k)$ 
    }
    
    Compute supervised loss $\mathcal{L}_{\text{Sup}}$ using $P_{\{s,c,a\}}$ and ${Y}_{\text{s/c/a}}$\;
    Compute consistency loss $\mathcal{L}_{\text{Unsup}}$ on $X^u$ across three views \;
    Update parameters minimizing $\mathcal{L}_{\text{total}} = (1-\beta)\mathcal{L}_{\text{Sup}} + \beta\mathcal{L}_{\text{Unsup}}$\;
}
\end{algorithm*}

\begin{figure*}[h]
    \centering
    \includegraphics[width=1.0\textwidth]{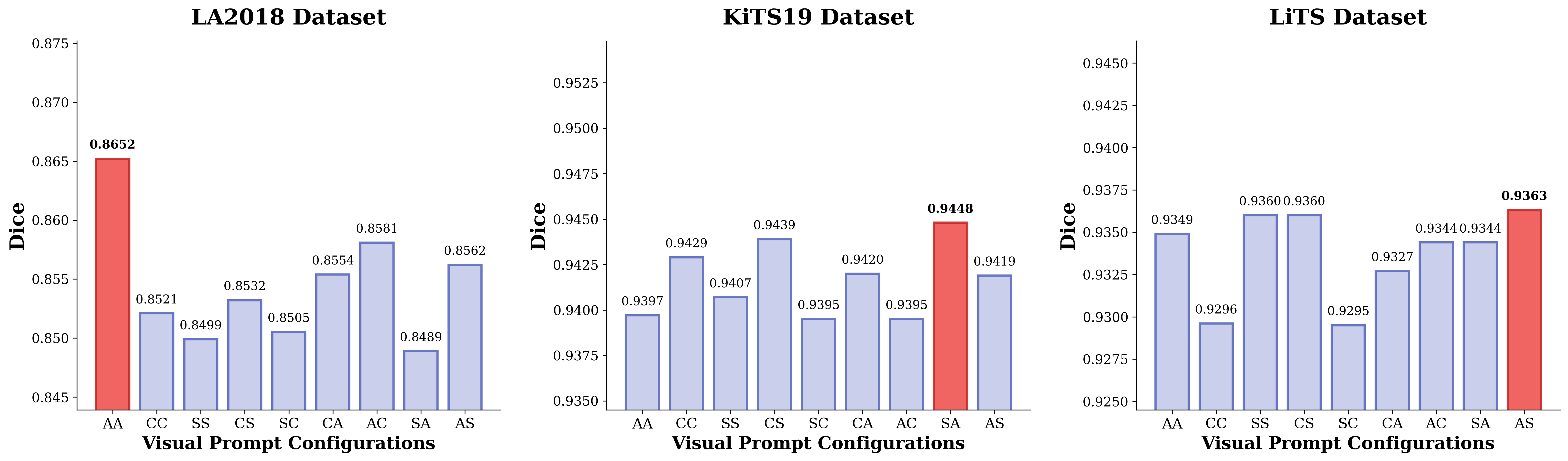} 
    \caption{Ablation study of visual prompt configurations on LA2018, KiTS19, and LiTS datasets under 10\% labeled cases. Red bars denote the optimal view strategy for each dataset.}
    \label{fig:view_ablation}
\end{figure*}

\subsection{Ablation Study on Visual Prompt Configurations}
\label{sec:ablation_views}

To further investigate the optimal setting for the visual prompt of Region-Aware Dual-encoder, we conduct an ablation study on the different input views for the auxiliary image embeddings. Our proposed RADA framework leverages a triple-view training scheme for the segmentation backbone. Inspired by this triple-branch design, we initially attempt to input all three slices into the image encoder. However, this doesn't get optimal results. Consequently, we adopt an approach of inputting two slices into the image encoder, as the visual encoder may exhibit varying sensitivities to different anatomical planes depending on the target organ. We evaluate various view configurations for the input of the image encoder while keeping the labeled slices for the segmentation network fixed to three orthogonal slices.
As illustrated in Figure \ref{fig:view_ablation},
the results demonstrate that visual prompt configurations combining orthogonal planes ({\itshape e.g.}, SA for KiTS19 with 0.9448 Dice, AS for LiTS with 0.9363 Dice) generally yield superior performance, as they effectively leverage complementary information from different anatomical perspectives. This orthogonal integration enhances the model's ability to capture comprehensive spatial relationships.
However, the LA2018 dataset achieves optimal performance with the AA configuration (0.8652 Dice), which can be attributed to the inherent characteristics of cardiac imaging. In LA2018, the axial plane contains the highest-resolution 2D slices with the most abundant supervisory signals, making axial-axial pairing particularly effective for left atrium segmentation. This exception underscores the importance of adapting visual prompt strategies to both anatomical specificity and data quality considerations.

\subsection{Visualization of Grad-CAM Heatmaps}
\begin{figure*}[t]
  \centering
  \includegraphics[width=0.9\textwidth]{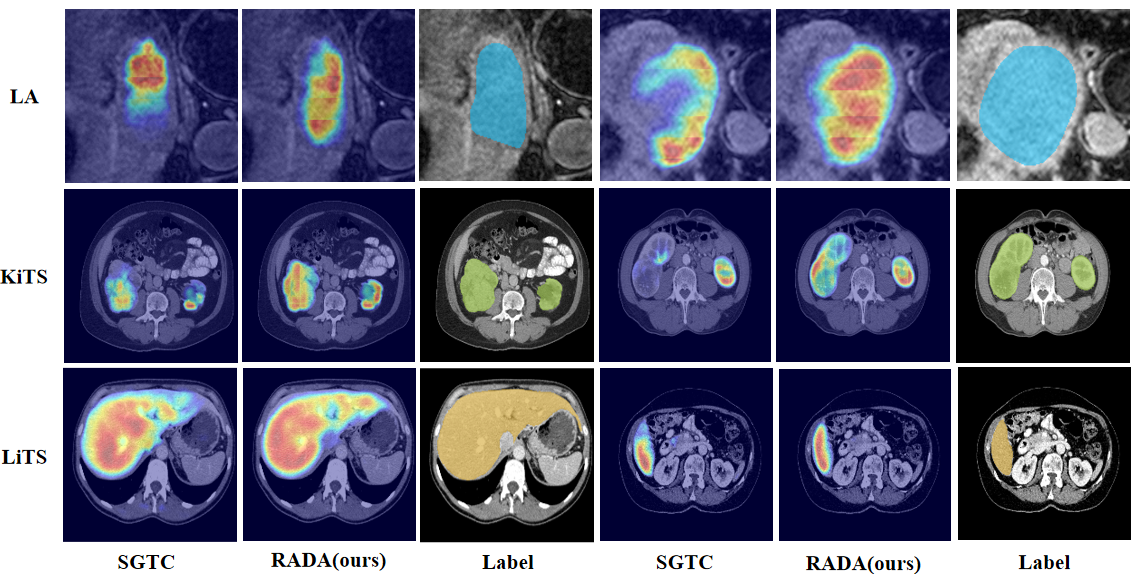}
  \caption{Visualization of Grad-CAM heatmaps on LA2018, KiTS19, LiTS dataset.}
  \label{heatmap}
\end{figure*}

To further demonstrate the effectiveness of our proposed RADA, we have generated heatmaps with Grad-CAM for LA2018, KiTS19, and LiTS dataset, as shown in Figure \ref{heatmap} respectively. In these visualizations, deeper colors represent higher activation levels, with red areas indicating regions of high confidence. The baseline SGTC exhibits scattered and weak activations that fail to cover the target organs completely. In contrast, our proposed RADA produces intense red activation regions that correspond more precisely with the labels. This confirms that RADA successfully captures fine-grained anatomical details and maintains higher confidence within the regions of interest.

\subsection{More Visualization Results}

\begin{table*}
\centering
\caption{A comprehensive ablation study evaluating the impact of different slice selection strategies on segmentation performance of LA2018 dataset under 10\% annotations.}
\label{LA_table}
\setlength{\tabcolsep}{10pt}
\scriptsize
\begin{tabular}{cc cccc} 
\toprule
\multirow{2}{*}{\textbf{\makecell{Labeled \\ Slices}}} & \multirow{2}{*}{\textbf{\makecell{Image Encoder \\ Slices}}} & \multicolumn{4}{c}{\textbf{Metrics}} \\
\cmidrule(lr){3-6} 
& & \textbf{Dice $\uparrow$} & \textbf{Jaccard $\uparrow$} & \textbf{HD $\downarrow$} & \textbf{ASD $\downarrow$} \\
\toprule

SSS & SS & 0.7014 $\pm$ 0.0950 & 0.5479 $\pm$ 0.1134 & 38.35 $\pm$ 9.79 & 13.18 $\pm$ 4.70 \\
\cmidrule{1-6}
CCC & CC & 0.7387 $\pm$ 0.0692 & 0.5900 $\pm$ 0.0843 & 29.37 $\pm$ 11.67 & 9.76 $\pm$ 4.12 \\
\cmidrule{1-6}
AAA & AA & 0.7937 $\pm$ 0.0473 & 0.6604 $\pm$ 0.0644 & 27.63 $\pm$ 7.92 & 4.66 $\pm$ 4.27 \\
\toprule

\multirow{3}{*}{ACA} & AA & 0.7859 $\pm$ 0.0492 & 0.6500 $\pm$ 0.0675 & 30.69 $\pm$ 10.43 & 5.91 $\pm$ 4.66 \\
 & CC & 0.7909 $\pm$ 0.0448 & 0.6562 $\pm$ 0.0601 & \textbf{28.08 $\pm$ 9.33} & \textbf{4.09 $\pm$ 2.75} \\
 & AC & \textbf{0.7985 $\pm$ 0.0451} & \textbf{0.6668 $\pm$ 0.0622} & {30.13 $\pm$ 9.59} & {5.76 $\pm$ 4.37} \\
\cmidrule{1-6}

\multirow{3}{*}{ASA} & AA & 0.8072 $\pm$ 0.0374 & 0.6783 $\pm$ 0.0536 & 31.05 $\pm$ 12.81 & 6.34 $\pm$ 4.46 \\
 & SS & 0.8087 $\pm$ 0.0390 & 0.6805 $\pm$ 0.0564 & 24.09 $\pm$ 9.14 & 4.79 $\pm$ 3.02 \\
 & AS & \textbf{0.8208 $\pm$ 0.0403} & \textbf{0.6980 $\pm$ 0.0588} & \textbf{22.73 $\pm$ 9.12} & \textbf{4.79 $\pm$ 2.76} \\
\cmidrule{1-6}

\multirow{3}{*}{CAC} & AA & 0.8246 $\pm$ 0.0310 & 0.7026 $\pm$ 0.0448 & 16.67 $\pm$ 3.66 & 3.67 $\pm$ 1.42 \\
 & CC & 0.8136 $\pm$ 0.0363 & 0.6873 $\pm$ 0.0514 & 19.05 $\pm$ 8.65 & 4.28 $\pm$ 1.72 \\
 & AC & \textbf{0.8289 $\pm$ 0.0316} & \textbf{0.7089 $\pm$ 0.0460} & \textbf{15.75 $\pm$ 3.61} & \textbf{3.23 $\pm$ 1.15} \\
\cmidrule{1-6}

\multirow{3}{*}{CSC} & CC & 0.8412 $\pm$ 0.0347 & 0.7274 $\pm$ 0.0514 & 15.90 $\pm$ 4.56 & 3.02 $\pm$ 1.63 \\
 & \cellcolor{gray!20}SS & \cellcolor{gray!20}\textbf{0.8515 $\pm$ 0.0284} & \cellcolor{gray!20}\textbf{0.7424 $\pm$ 0.0436} & \cellcolor{gray!20}\textbf{15.39 $\pm$ 4.60} & \cellcolor{gray!20}\textbf{2.87 $\pm$ 1.54} \\
 & SC & 0.8412 $\pm$ 0.0347 & 0.7274 $\pm$ 0.0514 & 15.90 $\pm$ 4.56 & 3.02 $\pm$ 1.63 \\
\cmidrule{1-6}

\multirow{3}{*}{SAS} & SS & 0.7642 $\pm$ 0.0624 & 0.6224 $\pm$ 0.0826 & 33.65 $\pm$ 9.35 & 10.42 $\pm$ 3.43 \\
 & AA & 0.7617 $\pm$ 0.0772 & 0.6211 $\pm$ 0.1006 & 31.99 $\pm$ 12.25 & 9.77 $\pm$ 4.52 \\
 & AS & \textbf{0.7722 $\pm$ 0.0705} & \textbf{0.6340 $\pm$ 0.0929} & \textbf{31.04 $\pm$ 10.05} & \textbf{9.11 $\pm$ 3.97} \\
\cmidrule{1-6}

\multirow{3}{*}{SCS} & SS & 0.8072 $\pm$ 0.0374 & 0.6783 $\pm$ 0.0536 & 31.05 $\pm$ 12.81 & 6.34 $\pm$ 4.46 \\
 & CC & 0.8087 $\pm$ 0.0390 & 0.6805 $\pm$ 0.0564 & 24.09 $\pm$ 9.14 & 4.79 $\pm$ 3.02 \\
 & SC & \textbf{0.8208 $\pm$ 0.0403} & \textbf{0.6980 $\pm$ 0.0588} & \textbf{22.73 $\pm$ 9.12} & \textbf{4.39 $\pm$ 2.76} \\

\toprule

SCA & AA & \textbf{0.8652 $\pm$ 0.0282} & \textbf{0.7634 $\pm$ 0.0441} & \textbf{15.17 $\pm$ 10.90} & \textbf{3.24 $\pm$ 2.23} \\
\bottomrule
\end{tabular}
\end{table*}

\begin{figure*}[t]
  \centering
  \includegraphics[width=0.9\textwidth]{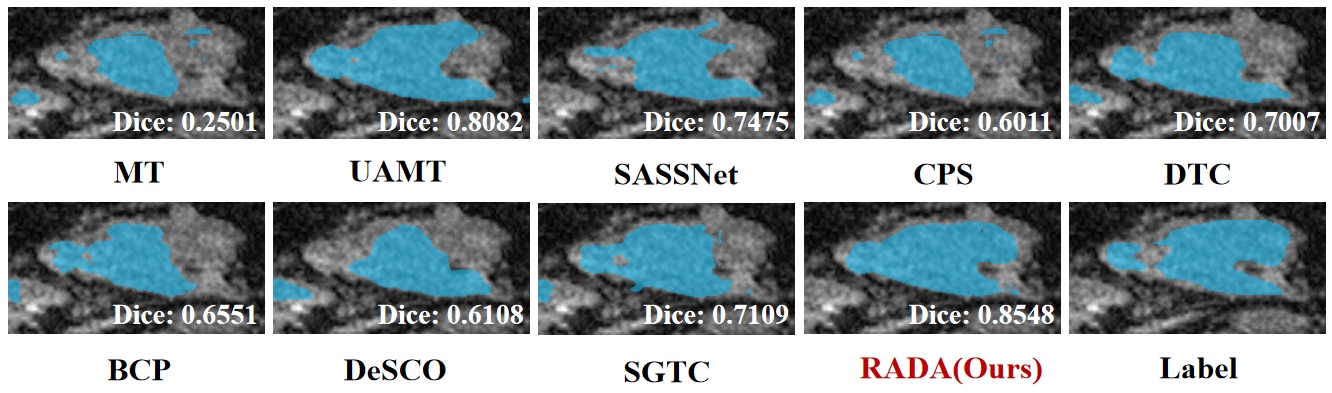}
  \caption{Quantitative comparisons on LA2018 dataset with 10\% labeled case.}
  \label{LA_seg}
\end{figure*}
\noindent \textbf{Visual Comparisons on LA2018 Dataset:} 
We further provide the qualitative comparisons on the LA2018 dataset in Figure \ref{LA_seg}. As observed, the left atrium possesses irregular and complex boundaries, which poses a significant challenge for baseline methods. Methods like MT and CPS suffer from severe under-segmentation and fail to capture the complete anatomical structure, resulting in disconnected fragments. In contrast, our RADA accurately delineates the intricate boundaries and preserves the structural integrity of the left atrium. By leveraging region-aware visual features, RADA effectively suppresses false negatives and generates segmentation masks that are more consistent with the label, achieving the highest Dice score of 0.8548 in this sample.

\begin{figure*}[t]
  \centering
  \includegraphics[width=0.9\textwidth]{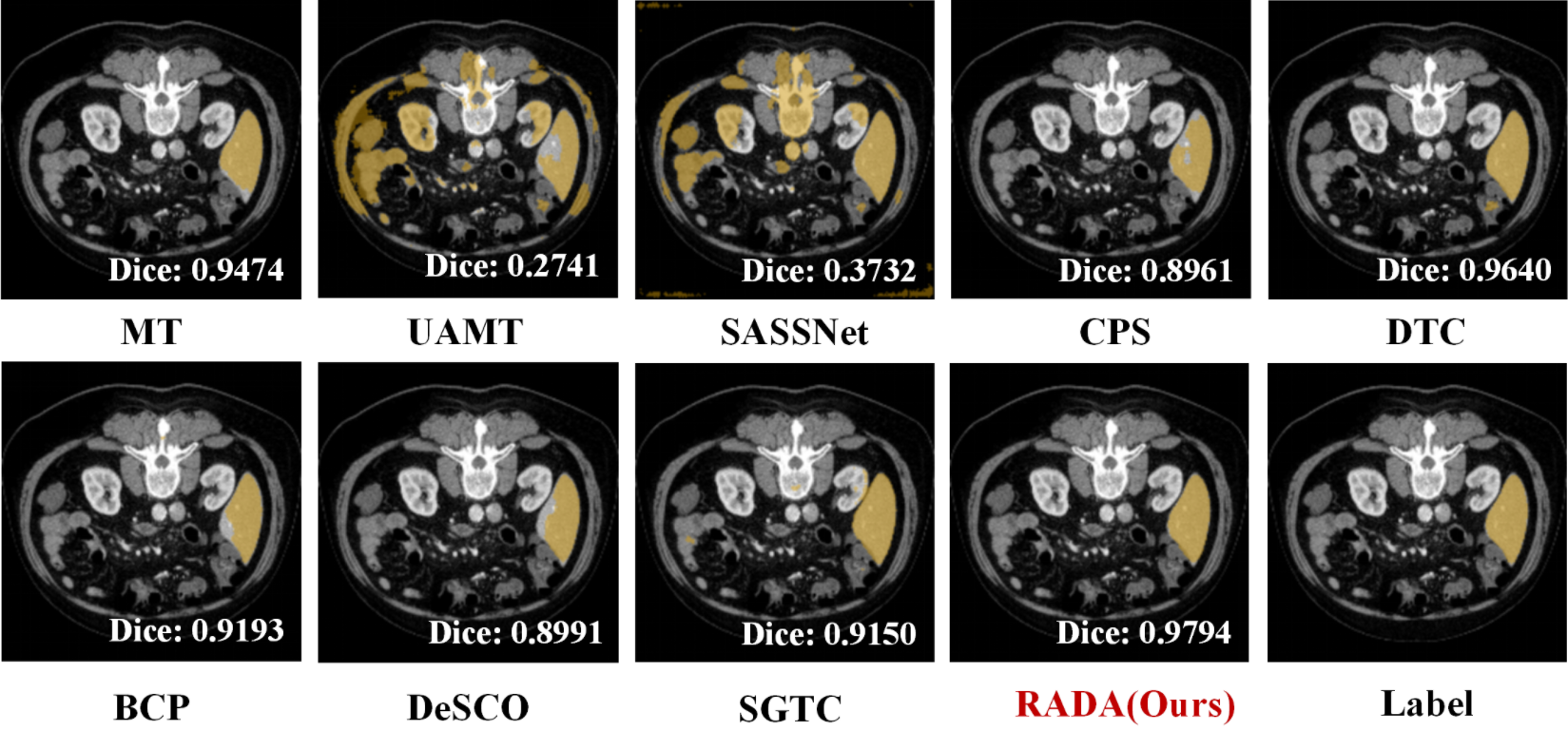}
  \caption{Quantitative comparisons on LiTS dataset with 10\% labeled case.}
  \label{LiTS_seg}
\end{figure*}

\noindent \textbf{Visual Comparisons on LiTS Dataset:} 
We also provide qualitative comparisons on LiTS dataset in Figure \ref{LiTS_seg}. While semi-supervised methods such as UAMT and SASSNet struggle with inaccurate boundary localization and incomplete lesion detection -- resulting in low Dice scores of 0.2741 and 0.3732, respectively -- our proposed RADA demonstrates superior performance in capturing fine structural details and maintaining region consistency. Notably, RADA achieves a Dice score of 0.9794, effectively suppressing both false positives and undersegmentation, and produces segmentation masks that align closely with the label. This result highlights the ability of our method to leverage limited annotated data while retaining robustness against complex anatomical variations.

\subsection{Ablation Study on Different Slice Selection Strategies}

Based on the comprehensive experimental results in Table \ref{LA_table}, our analysis reveals clear performance patterns that demonstrate the critical importance of orthogonal slice selection in the RADA framework. The data establishes a definitive performance hierarchy where orthogonal slice combinations consistently outperform parallel configurations, with the SCA strategy (incorporating sagittal, coronal, and axial views) achieving optimal performance at a Dice score of 0.8652. This represents a significant improvement over both parallel slice arrangements (SSS, CCC, AAA achieving 0.7014 - 0.7937 Dice) and partial orthogonal configurations (SCS reaching 0.8208 Dice), confirming that each additional orthogonal plane contributes valuable complementary anatomical information. The superiority of orthogonal approaches stems from their ability to capture comprehensive spatial relationships from multiple viewing perspectives, enabling more robust 3D understanding compared to the limited viewpoint consistency provided by parallel slices. Furthermore, within orthogonal training frameworks, the strategic selection of orthogonal visual prompt pairs (such as AS combinations outperforming AA/SS pairs in ASA configurations) provides additional complementary spatial information that enhances anatomical understanding. These findings have crucial practical implications for barely-supervised learning paradigms, demonstrating that when annotation resources are limited, prioritizing orthogonal diversity over quantity can maximize segmentation performance. The success of our SCA approach confirms that with just three strategically chosen orthogonal slices, we can achieve performance comparable to more heavily annotated methods, establishing orthogonal integration as a fundamental principle for efficient medical image segmentation in clinical applications where annotation resources are constrained.

\end{document}